\documentclass[11pt]{article}

\usepackage[final]{acl}

\usepackage{times}
\usepackage{latexsym}

\usepackage[T1]{fontenc}
\usepackage[table]{xcolor}
\usepackage{amsmath} 
\usepackage[utf8]{inputenc}

\usepackage{microtype}

\usepackage{graphicx}

\usepackage{booktabs}
\usepackage{multirow}
\usepackage{graphicx}
\usepackage[table]{xcolor}
\usepackage{makecell}

\usepackage{algorithm}
\usepackage{algpseudocode}

\usepackage{amsmath}
\usepackage{amssymb}

\usepackage{booktabs}

\title{KARMA: Knowledge graph-based Automated Reasoning Materialization and Alignment}

\author{
  Jinkyeong Choi, Chaebin Jeong, and
  Donghyeon Park\thanks{Corresponding author.} \\
  Sejong University \\
  Seoul, Republic of Korea \\
  \texttt{jjinchoi@sju.ac.kr} \quad
  \texttt{chaibin2000@sju.ac.kr} \quad
  \texttt{parkdh@sejong.ac.kr}
}

\begin{document}
\maketitle
\begin{abstract}
Template-based contrastive synthesis is scalable, but its candidates often differ only in a few entity-slots while sequence-level optimization spreads supervision over mostly shared templates. We formalize this as the Resolution Mismatch Problem and propose KARMA, which enumerates schema-constrained paths over domain knowledge graphs and verbalizes them into slot-aligned contrastive candidates. Slot-Parallel Alignment (SPA) then applies a decoupled slot-level objective to route preference supervision to discriminative entity-slots, with slot-aware masked attention serving as an optional packed-evaluation implementation. Across biomedical, computer-science, and chemistry benchmarks, KARMA outperforms base LLM and same-data SFT baselines, and compares favorably with sequence- and token-level preference methods.

\end{abstract}

\newcommand{\red}[1]{\textcolor{black}{#1}}
\newcommand{\blue}[1]{\textcolor{black}{#1}}

\section{Introduction}

Recent progress in LLM reasoning has increased the demand for reasoning supervision~\cite{cot, deepseek-R1, jaech2024openai}. \red{However}, scaling such supervision remains bottlenecked by trade-offs among reliability, diversity, and cost~\cite{uesato2022solving, reasoning-trajectories}.

Prior approaches address this trade-off in three ways. 
(1) Human expert annotation yields reliable reasoning chains but is prohibitively expensive at the process level~\cite{uesato2022solving, reasoning-trajectories, math-shepherd}.
(2) Model-generated synthetic data improves scalability~\cite{self-instruct, wizardlm}, but can inherit source-model errors and degrade under recursive use~\cite{gudibande2023false, shumailov2024ai}.
(3) Rule- or template-based generation reduces cost through programmatic construction, but can restrict data diversity~\cite{riaz-etal-2025-metasynth}.
Despite these differences in how reasoning data is generated, a structural limitation emerges once it is used for contrastive supervision: candidate differences are often concentrated in a few entity-slots, while supervision is applied at the full-sequence level.

\begin{figure}[t]
  \includegraphics[width=\columnwidth]{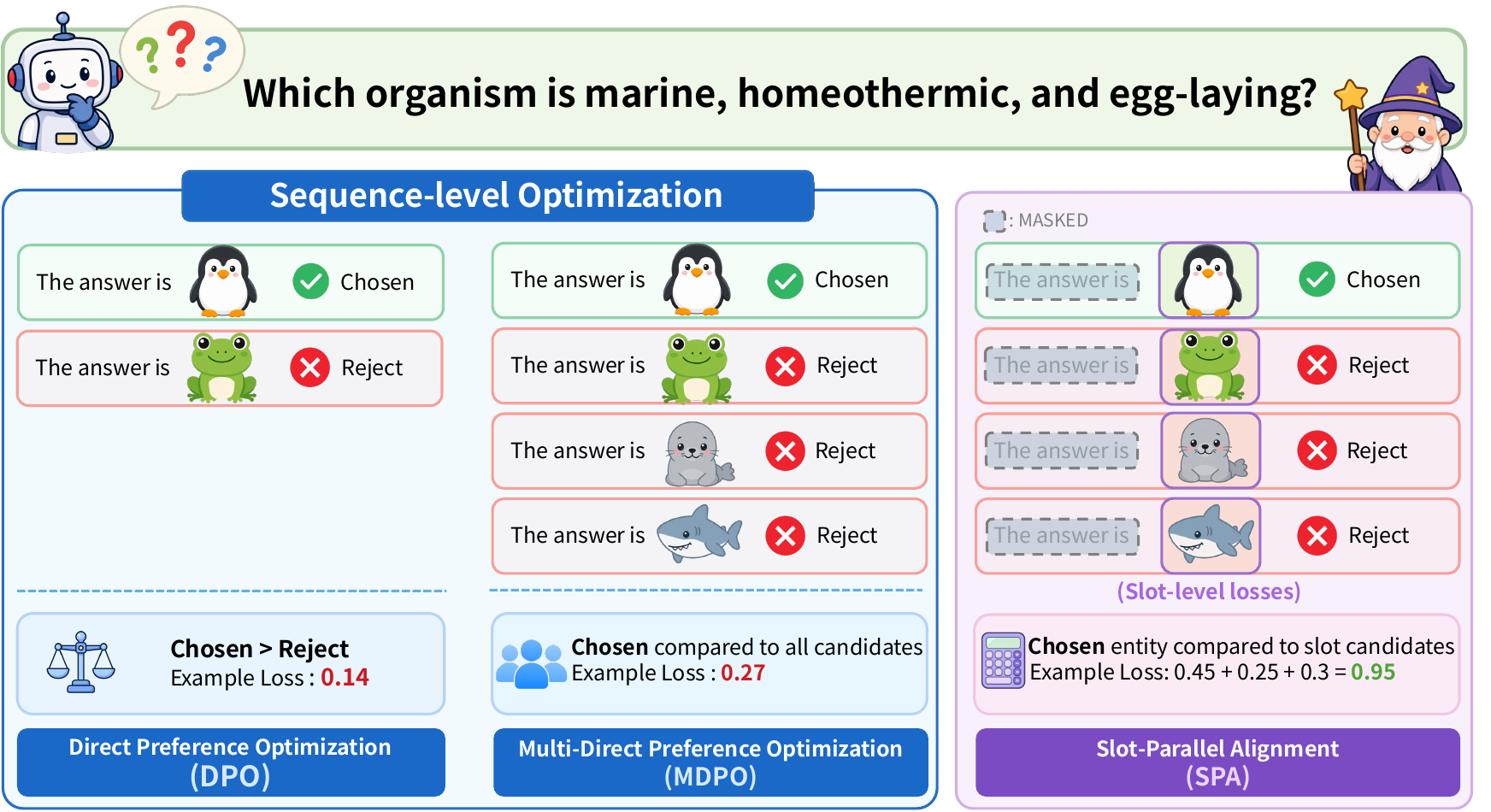}
    \caption{The Resolution Mismatch Problem in preference optimization. 
    (a) DPO compares a single chosen-rejected pair at the sequence level. 
    (b) Multi-DPO extends this to multiple rejected candidates, still at the sequence level. 
    (c) Our Slot-Parallel Alignment (SPA) compares candidates sharing the same template and aligns preference at the resolution of discriminative entity-slots. 
    The loss values illustrate supervision granularity and are not numerically comparable across objectives.}  \label{fig:figure1}
\end{figure}

We formalize this gap as the \textit{Resolution Mismatch Problem} (Fig.~\ref{fig:figure1}): preference is often determined by a few discriminative entity-slots, while optimization is applied to the full sequence. \textit{The bottleneck is not how many candidates we compare, but at what resolution we compare them.} Standard DPO~\cite{dpo} (Fig.~\ref{fig:figure1}-a) and its multi-candidate extensions (Fig.~\ref{fig:figure1}-b) widen the comparison set but retain sequence-level granularity~\cite{PRO, LiPO}, which can dilute the preference signal associated with informative entity-slots.
We examine this behavior directly through gradient analysis, measuring how sequence-, token-, and slot-level objectives distribute sensitivity across entity and template tokens.

Knowledge graphs provide a natural way to address this mismatch. 
Prior work uses KG paths as structured intermediate traces for reasoning~\cite{RoG, ToG, PoG}. 
We use this path structure differently for contrastive data construction: under fixed schemas, alternative paths are verbalized with a shared template, with differences localized to entity-slots.
We instantiate this idea in \textbf{KARMA} (\textbf{K}nowledge graph-based \textbf{A}utomated \textbf{R}easoning \textbf{M}aterialization and \textbf{A}lignment), a unified framework whose synthesis pipeline produces such candidates and whose learning component, \textbf{Slot-Parallel Alignment (SPA)}, operates directly on them (Fig.~\ref{fig:figure1}-c).
KARMA uses the KG only at data-construction time; the trained model is evaluated without graph access, making inference-time KG grounding methods complementary rather than direct alternatives.
Prior KG-to-text methods verbalize structured facts into natural language~\cite{WebNLG, KELM}. 
The synthesis pipeline verbalizes selected KG paths under a shared template and uses recurrence-based support as a structural prior for preference assignment. \red{This exposes} explicit discriminative entity-slots while expanding factual coverage beyond hand-written templates.

On the learning side, SPA applies a decoupled slot-level objective that supervises construction-exposed entity slots while retaining the shared template as a language-modeling signal. For efficient packed evaluation, SPA can optionally use slot-aware masked attention to approximate per-candidate log-likelihoods in a single forward pass without changing the slot-level supervision signal.
Unlike prior token-level methods that identify, estimate, or learn token importance~\cite{cDPO, TIS-DPO, sepo, SparsePO, TI-DPO}, SPA targets entity-slots that are explicitly exposed by construction in KARMA-synthesized data.

When data and optimization share the same slot-level resolution, preference learning can focus on discriminative evidence while preserving shared templates. We evaluate KARMA across biomedical, chemistry, and computer-science benchmarks, where it improves over base LLM and SFT baselines on most benchmarks and compares favorably with sequence- and token-level preference methods.
Our contributions are:
\begin{itemize}
\item \textbf{Aligned contrastive synthesis from KGs:} We propose a KG-grounded synthesis pipeline that constructs chosen-rejected candidates sharing a verbalization template by design, localizing preference differences to entity-slots and using KG path recurrence as a structural prior for preference assignment.
\item \textbf{Slot-Parallel Alignment (SPA):} We introduce a preference optimization method whose resolution matches the data: a decoupled slot-level objective applies preference and anchor losses to construction-exposed entity slots while preserving shared templates as a language-modeling signal. We further provide an optional slot-aware masked attention implementation that reduces redundant candidate evaluation without estimating token importance.
\end{itemize}

\section{Related Work}

\paragraph{Trade-offs in Reasoning Supervision}
Reasoning supervision faces a cost-coverage trade-off: human process supervision yields reliable step-level signals but is prohibitively expensive~\cite{uesato2022solving, reasoning-trajectories, math-shepherd}, model-generated data scales better~\cite{self-instruct, wizardmath} but can inherit source-model errors and degrade under recursive use~\cite{gudibande2023false, shumailov2024ai}, and template-based generation lowers cost but can restrict data diversity~\cite{riaz-etal-2025-metasynth}.
When such data is used for contrastive supervision, candidate differences can become concentrated in a few entity-slots.
KARMA turns this localized variation into a resource: by grounding candidate construction in KG paths, it expands factual coverage while keeping discriminative variation localized to entity-slots by construction.

\paragraph{Fine-grained Preference Alignment}
Sequence-level preference optimization applies supervision to sequence-level scores, including pairwise objectives~\cite{dpo, bdpo} and ranking-based objectives over multiple candidates~\cite{PRO, LiPO}.
When candidates share most tokens and differ only at a few positions, this granularity does not explicitly localize supervision to the discriminative positions.
Token-level methods make preference optimization more fine-grained through token-wise KL or reward regularization~\cite{TDPO, T-REG}, while other methods identify or weight important tokens using contrastive estimation, importance sampling, reward estimation, or learned masks~\cite{cDPO, TIS-DPO, sepo, SparsePO, TI-DPO}.
SPA instead exploits a structural prior from KARMA-synthesized data: discriminative entity-slots are known by construction, so supervision routes to them directly without estimation.

\paragraph{Knowledge Graph Verbalization}
Prior work verbalizes structured KG facts into natural language for data generation and language-model adaptation~\cite{WebNLG, KELM}.
KG-augmented reasoning methods instead use graph paths or graph-derived plans to support retrieval and reasoning~\cite{LPKG, RoG, ToG, PoG}.
KARMA uses this structure differently: rather than supplying graph evidence at inference, it constructs multiple training candidates that share a source-target pair, relation schema, and verbalization template, exposing intermediate entity variations as slot-level contrastive supervision.

\section{Methodology}

\subsection{Overview}
We propose \textbf{KARMA}, a knowledge graph (KG)-grounded contrastive synthesis pipeline, and \textbf{SPA}, a slot-level preference optimization method co-designed for KARMA-synthesized data. KARMA enumerates alternative KG paths between fixed source-target entities under a fixed relational schema and verbalizes them into structurally aligned contrastive candidates that share a template scaffold while differing only at discriminative entity-slots. SPA then aligns preference optimization with this entity-slot-level structure through two core learning mechanisms: (i) slot-parallel organization of candidate variants under shared context, and (ii) a decoupled hybrid objective that applies preference supervision to entity slots while preserving the shared template. Slot-aware masked attention is an optional packed-evaluation mechanism that reduces redundant forward passes by approximating per-candidate log-likelihoods in a single forward pass. Together, KARMA and SPA realize a single principle: \textbf{align the granularity of supervision with the structural granularity of the data}.

\begin{figure*}[h]
    \centering
    \includegraphics[width=0.9\linewidth]{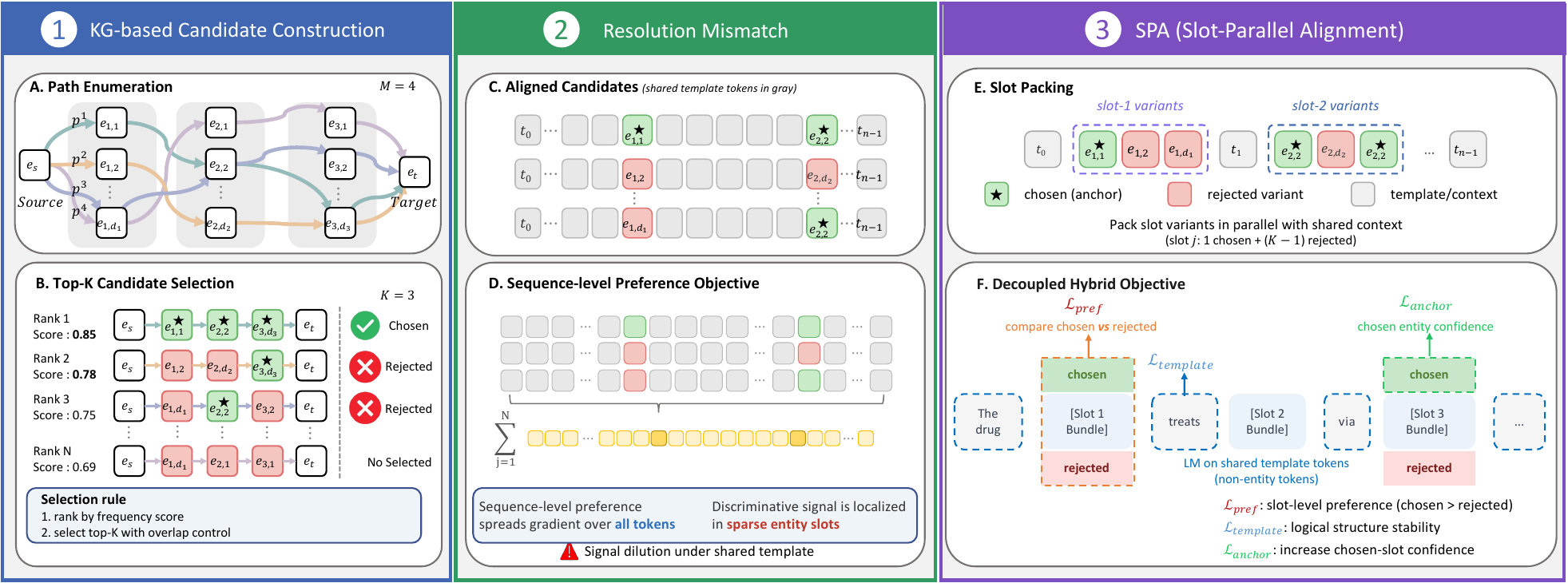}
    \caption{Overview of KARMA and SPA. KARMA synthesizes structurally aligned contrastive candidates from knowledge-graph paths via path enumeration, support-based top-$K$ selection, and template verbalization. Because candidate differences are localized to sparse entity-slots, sequence-level preference optimization can suffer from signal dilution. SPA addresses this mismatch with a decoupled slot-level objective that aligns learning with entity-slot-level supervision. Slot-aware masked attention, when enabled, serves as an optional packed-evaluation implementation that reduces redundant candidate evaluation while leaving the slot-level supervision signal unchanged. The support values in Panel B are normalized within each candidate pool for visualization only; path selection uses the raw recurrence-based scores.}
    \label{fig:figure2}
\end{figure*}

\subsection{KARMA Data Synthesis: KG-Grounded Contrastive Candidate Construction}
\label{sec:karma}
KARMA constructs slot-aligned contrastive candidates from a domain
knowledge graph in three steps: it enumerates schema-constrained paths
between source-target entity pairs, selects top-$K$ alternatives per
candidate pool by balancing support and structural diversity, and
verbalizes the selected paths with a schema-specific template shared
within the pool. The resulting candidates share the same source and
target entities and relation schema, differing only at their
intermediate entity slots. This slot-localized variation is the
interface to SPA: it confines contrastive information to entity slots,
exposing preference signal at the same granularity as the data
variation.

\subsubsection{Path Enumeration}
\label{sec:path-enum}
KARMA first constructs candidate pools by enumerating schema-constrained multi-hop paths from a domain knowledge graph (KG) $\mathcal{G}=(\mathcal{N},\mathcal{E})$, where $\mathcal{N}$ is the set of entities and $\mathcal{E}\subseteq
\mathcal{N}\times\mathcal{R}\times\mathcal{N}$ is the set of typed edges over relation types $\mathcal{R}$. A relation schema specifies an ordered pattern of typed edges that candidate paths must instantiate. Schemas are exhaustively enumerated rather than selected for individual source-target pairs, as detailed in Appendix~\ref{app:relation_schemas}. KARMA uses such schemas to collect alternative paths that share the same schema and source and target entities while differing only in their intermediate entities (Fig.~\ref{fig:figure2}A).

For a given source entity $e_s$, target entity $e_t$, and relation schema, KARMA enumerates a schema-constrained candidate pool $\mathcal{P}$, \blue{a source-to-target subgraph of $\mathcal{G}$,} whose elements take the form
\begin{equation}
\blue{
p^{m} = \left( e_s, e_1^{(m)}, e_2^{(m)}, \ldots, e_{n-1}^{(m)}, e_t \right)
},
\end{equation}
where $m=1,\dots,M$ indexes paths within the pool, $n$ denotes the number of hops in the relation schema, $e^{(m)}_j$ denotes the entity at the $j$-th intermediate position of path $p^m$ for $j=1,\dots,n-1$, and $e_s,e_t$ denote the source and target entities.

Each such pool corresponds to one source-target pair and serves as the candidate set for one training instance. Paths in the same pool share the same relation schema and source and target entities, but may differ in their intermediate entities. \blue{Equivalently, each intermediate slot $j$ induces a set of distinct slot entities $\mathcal{S}_j=\{e_{j,1},\dots,e_{j,d_j}\}$, from which each path selects one entity, \red{so $e_j^{(m)}\in\mathcal{S}_j$.}}
This property allows KARMA to construct contrastive alternatives whose differences are localized to intermediate entity slots.

\subsubsection{Top-$K$ Candidate Selection}
\label{sec:topk}
Given a candidate pool $\mathcal{P}$ of size $M$, written $\mathcal{P}=\{p^1,\dots,p^M\}$, KARMA selects $K$ paths by balancing path support and structural diversity. We measure the support of each path by the cumulative recurrence of its intermediate entities within the pool: each intermediate entity contributes its occurrence count across $\mathcal{P}$, and the path's support is the sum of these contributions. Paths whose intermediate entities frequently recur thus receive higher support, providing a training-free evidence prior from repeated graph traversals within the same source-target pool. This recurrence-based support is not a correctness guarantee: the highest-scoring path need not be uniquely correct or preferable.

KARMA then greedily admits paths in descending order of support, subject to an overlap budget $\delta$ that limits the number of intermediate entities shared with previously admitted paths (Fig.~\ref{fig:figure2}B). This budget enforces structural diversity by preventing near-duplicate candidates. The procedure stops when $K$ paths have been admitted; pools with fewer than $K$ admissible paths are discarded.


We denote the admitted paths by $p^{\star}_k$ ($k=1,\dots,K$), indexed in admission order, \blue{with intermediate entities written as $e_j^{(k)}$ for slot $j$ of the $k$-th selected path.} 
The ordered selected
path sequence
\begin{equation}
\mathcal{P}^{\star}=(p^{\star}_1,\dots,p^{\star}_K)
\end{equation}
preserves this ordering through verbalization and provides the preference ordering used by SPA: $p^{\star}_1$ corresponds to the chosen candidate and $p^{\star}_{2:K}$ to the rejected candidates.
This ordering represents relative evidence strength rather than a binary partition of path correctness: rejected candidates may remain valid at shared slots while differing at entity-slots with weaker support. This distinction motivates SPA's slot-level objective, which contrasts discriminative entity-slots without assigning a uniform negative signal to the shared template.
We empirically validate this chosen-rejected partition with a GPT-4 judge study in Appendix~\ref{app:chosen_validation}, and report end-to-end ablations against random path selection in Section~\ref{sec:abl}.

\subsubsection{Template Verbalization}
\label{sec:verbalization}
Each selected path $p^{\star}_k\in\mathcal{P}^{\star}$ is verbalized into a natural-language sequence $y^{k}$ by populating a schema-specific template with the entities along the path. Verbalization is fully rule-based and uses \textit{no language model in the generation loop}. For each schema, we predefine multiple template variants that differ in wording and synonym choice and uniformly sample one variant for each candidate pool. The sampled template provides a fixed scaffold of textual fragments and entity slots determined by the relation schema, while the entities of $p^{\star}_k$ are realized as multi-token spans following a domain-specific rule.

Within a candidate pool, all selected paths share the same source and target entities and the same relation schema, so they are verbalized under an identical template scaffold. Consequently, the resulting ordered candidate sequence $\mathcal{Y}=(y^{1},\dots,y^{K})$ differs only at the intermediate entity slots, while all template fragments and endpoint entities remain shared (Fig.~\ref{fig:figure2}C). This slot-localized variation is precisely the structural property SPA exploits in the next section. Holding the sampled variant fixed within each pool preserves slot alignment, while sampling across the variant set introduces surface-form diversity across pools.

Repeating this procedure over all retained source-target pairs yields the training corpus
\begin{equation}
\mathcal{D}=\{(x_i,\mathcal{Y}_i)\}_{i=1}^{N},
\qquad
\mathcal{Y}_i=(y_i^{1},\dots,y_i^{K}),
\end{equation}
where $x_i$ is a prompt constructed from the source and target entities of the $i$-th instance, and $\mathcal{Y}_i$ is the ordered verbalized candidate set inheriting the chosen/rejected ordering from the corresponding selected path sequence.

\subsection{Slot-Parallel Alignment for Slot-Level Preference Learning}
\label{sec:spa}
SPA realigns preference optimization from sequence-level to slot-level resolution, matching the granularity at which KARMA candidates actually differ. It concentrates supervision on discriminative entity slots through a decoupled hybrid objective: slot-level preference and anchor losses supervise entity tokens, while a template LM loss preserves the shared scaffold. Slot-parallel packing evaluates variants under shared context, and slot-aware masked attention optionally obtains per-candidate log-likelihoods in one forward pass instead of $K{+}1$, at a small accuracy cost.

\subsubsection{Slot-Parallel Packing with Shared Context}
For a single training instance, we drop the corpus index $i$ and
write $\mathcal{Y}=(y^{1},\dots,y^{K})$. Following the path notation
in Section~\ref{sec:topk}, the candidates in $\mathcal{Y}$
differ only at the intermediate slots $j=1,\dots,n-1$, where slot $j$ holds the entity $e_j^{(k)}$ in candidate $y^{k}$.
Let $t_0,\dots,t_{n-1}$ be the shared template fragments, and let
$e_j^{(1)}$ denote the chosen entity at slot $j$, which we refer to
as the \emph{anchor}. SPA constructs the packed sequence
\begin{equation}
\mathbf{s}
=
\bigl[
x;\;
t_0,\; e_1^{(1{:}K)},\; t_1,\;
\dots,\; e_{n-1}^{(1{:}K)},\; t_{n-1}
\bigr],
\end{equation}
where $e_j^{(1{:}K)}$ denotes the ordered concatenation of slot-$j$
entity variants $e_j^{(1)},\dots,e_j^{(K)}$. As illustrated in
Fig.~\ref{fig:figure2}E, SPA groups the variants of each
intermediate slot into a local slot bundle while sharing the
surrounding template fragments. We shuffle the intra-slot variant
order during training to prevent positional shortcuts.

\begin{figure}[t]
  \includegraphics[width=\columnwidth]{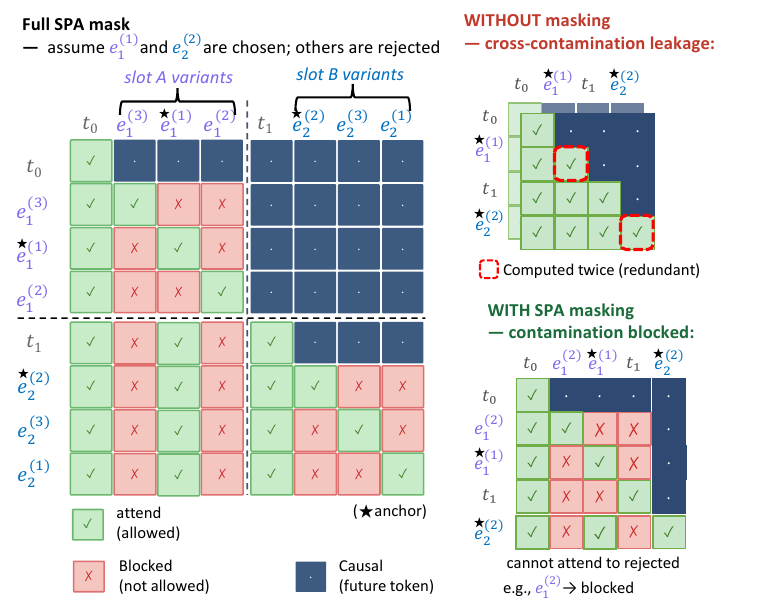}
  \caption{Slot-aware masked attention for packed candidate evaluation. Green cells denote allowed attention, red cells denote blocked attention, and dark cells denote future tokens. Compared with a plain causal mask, the SPA mask lets each variant attend to the shared template scaffold and its own slot state, while restricting previous slots to the anchored prefix \red{so that each candidate's log-likelihood within the packed sequence approximates its standalone value under $K{+}1$ independent forward passes}.}
  \label{fig:figure3}
\end{figure}

\subsubsection{Slot-Aware Masked Attention}
Since candidates in $\mathbf{s}$ share all template fragments and differ only at the intermediate entity slots, $K{+}1$ standalone forward passes would redundantly recompute identical template representations. SPA replaces the causal attention mask with a slot-aware mask so that all per-candidate log-likelihoods are obtained from a single forward pass while approximating their standalone values.

We organize the mask by what each query token is allowed to attend to within the causal prefix. Template tokens at fragment $t_j$ attend only to the preceding template fragments and to the \emph{anchor} entity of each past slot. An entity token $e_j^{(k)}$ at slot $j$ in candidate $k$ attends to the preceding template fragments, to the past anchors $e_1^{(1)},\dots,e_{j-1}^{(1)}$, and to itself. \red{Consequently, each candidate's log-likelihood within the packed sequence is computed along its own anchored path, approximating the value it would have under an independent forward pass.}

Fig.~\ref{fig:figure3} (left) illustrates the resulting mask for a packed sequence with two intermediate slots and three variants per slot. With a plain causal mask (top right), later tokens can attend to multiple variants from previous slots, \red{so a candidate's log-likelihood within the packed sequence diverges from its standalone value}. The SPA mask (bottom right) restricts each candidate to its own anchored path while keeping the template representation shared, so one forward pass over $\mathbf{s}$ \red{approximates $K{+}1$ independent passes} while encoding the template scaffold only once.

\subsubsection{Decoupled Hybrid Objective}
\red{Given per-candidate log-likelihoods obtained from the packed sequence, applying} a sequence-level preference objective to $\mathbf{s}$ would still spread supervision over many template tokens that are identical across candidates. Since KARMA candidates differ only at intermediate entity slots, SPA decouples the objective according to this structure: a
slot-level preference loss supervises entity choices, an anchor loss stabilizes the chosen entities, and a template language modeling loss preserves the shared scaffold (Fig.~\ref{fig:figure2}F).
\paragraph{Slot-level preference.}
For each slot $j$, the anchor entity $e_j^{(1)}$ should be preferred to the rejected entities $e_j^{(k)}$ from candidates $k=2,\dots,K$.
\blue{Let $\pi_\theta$ denote the policy model and $\pi_{\mathrm{ref}}$ a fixed reference model. For an entity span $e_j^{(k)}$, we write $\ell_\theta(e_j^{(k)}\mid\mathbf{s})$ and $\ell_{\mathrm{ref}}(e_j^{(k)}\mid\mathbf{s})$ for its length-normalized log-likelihood under $\pi_\theta$ and $\pi_{\mathrm{ref}}$, computed under the slot-aware mask.} We define the reference-normalized score
\begin{equation}
r_\theta(e_j^{(k)})
=
\ell_\theta(e_j^{(k)} \mid \mathbf{s})
-
\ell_{\mathrm{ref}}(e_j^{(k)} \mid \mathbf{s}).
\end{equation}
\blue{This score is the policy's log-likelihood gain over the reference on the entity span $e_j^{(k)}$.
}
SPA applies a squared margin loss between the anchor and each rejected entity at the same slot:
\begin{equation}
\label{eq:pref}
\begin{split}
\mathcal{L}_{\mathrm{pref}}
&= \frac{1}{(n-1)(K-1)} \\
&\quad \times \sum_{j=1}^{n-1} \sum_{k=2}^{K}
\left( r_\theta(e_j^{(1)}) - r_\theta(e_j^{(k)}) - \frac{1}{2\beta} \right)^2,
\end{split}
\end{equation}
where $\beta$ controls the target margin. Here, $k$ follows the original top-$K$ candidate order: $k=1$ is the chosen candidate and $k=2,\dots,K$ are rejected candidates. This term provides the
discriminative signal at the slots where candidates differ, rather than spreading preference supervision over the shared template.
\paragraph{Anchor confidence.}
The preference term in Eq.~\ref{eq:pref} constrains relative scores, but it does not directly
ensure that the chosen entity has high likelihood. SPA therefore adds an anchor confidence term:
\begin{equation}
\label{eq:anchor}
\mathcal{L}_{\mathrm{anchor}}
=
-\frac{1}{n-1}
\sum_{j=1}^{n-1}
\ell_\theta(e_j^{(1)} \mid \mathbf{s}).
\end{equation}
This term stabilizes the likelihood of the selected path entities,
complementing the relative comparison imposed by
$\mathcal{L}_{\mathrm{pref}}$.
\paragraph{Template language modeling.}
The template fragments encode the relation scaffold shared by all
candidates, but they are not discriminative for the slot-level
preference comparison. Let $\mathcal{T}$ denote the set of shared template token positions in $\mathbf{s}$\blue{, and let $s_u$ denote the $u$-th token of $\mathbf{s}$}. SPA preserves the scaffold with a standard causal language modeling loss:
\begin{equation}
\label{eq:template}
\mathcal{L}_{\mathrm{template}}
=
-\frac{1}{|\mathcal{T}|}
\sum_{u\in\mathcal{T}}
\log \pi_\theta(s_u \mid s_{<u}).
\end{equation}
This term maintains the fluency and logical structure of the
verbalization while leaving entity discrimination to the slot-level losses.
\paragraph{Full objective.}
The final SPA objective combines the slot-level preference term with the two auxiliary losses:
\begin{equation}
\label{eq:spa_full_objective}
\mathcal{L}_{\mathrm{SPA}}
=
\mathcal{L}_{\mathrm{pref}}
+
\lambda_{\mathrm{anchor}}\mathcal{L}_{\mathrm{anchor}}
+
\lambda_{\mathrm{template}}\mathcal{L}_{\mathrm{template}},
\end{equation}
where $\lambda_{\mathrm{anchor}}$ and $\lambda_{\mathrm{template}}$ are scalar loss weights. We average $\mathcal{L}_{\mathrm{SPA}}$ over training instances.

\definecolor{best}{RGB}{215,213,240}     
\definecolor{second}{RGB}{230,240,255}    

\begin{table*}[h]
\centering
\small
\resizebox{1.7\columnwidth}{!}{%
\begin{tabular}{llccccc}
\toprule
\multirow{2}{*}{\textbf{Domain}} & \multirow{2}{*}{\textbf{Metric}} & \multirow{2}{*}{\textbf{Baseline}} & \multicolumn{2}{c}{\textbf{Simple SFT}} & \multirow{2}{*}{\textbf{Ours (KARMA)}} \\
\cmidrule(lr){4-5}
 & & \textbf{(Base LLM)} & \textbf{(1) Open-Data} & \textbf{(2) KARMA-Data} & \\
\midrule
\multirow{5}{*}{Biomedical}
 & MedQA (Acc)        & 0.362          & \cellcolor{second}0.434 & 0.273 & \cellcolor{best}\textbf{0.541} \\
 & PubMedQA (Acc)     & 0.511 & \cellcolor{second}0.512 & 0.229 & \cellcolor{best}\textbf{0.522} \\
 & MMLU-Bio (Acc)     & 0.505          & \cellcolor{second}0.571 & 0.381 & \cellcolor{best}\textbf{0.702} \\
 & MMLU-Pro-Bio (Acc) & 0.230          & \cellcolor{second}0.324 & 0.292 & \cellcolor{best}\textbf{0.532} \\
 & GPQA-Bio (Acc)     & \cellcolor{best}\textbf{0.700} & 0.533 & 0.333 & \cellcolor{second}0.667 \\
\midrule
\multirow{3}{*}{Computer Science}
 & MMLU-CS (Acc)      & 0.651          & \cellcolor{second}0.655 & 0.451 & \cellcolor{best}\textbf{0.701} \\
 & MMLU-Pro-CS (Acc)  & 0.195          & \cellcolor{second}0.243 & 0.180 & \cellcolor{best}\textbf{0.256} \\
 & CSBench (Acc)      & 0.345          & \cellcolor{best}\textbf{0.506} & 0.372 & \cellcolor{second}0.495 \\
\midrule
\multirow{3}{*}{Chemistry}
 & MMLU-Chem (Acc)      & \cellcolor{second}0.500 & 0.495 & 0.373 & \cellcolor{best}\textbf{0.630} \\
 & MMLU-Pro-Chem (Acc)  & 0.148          & \cellcolor{second}0.178 & 0.126 & \cellcolor{best}\textbf{0.195} \\
 & GPQA-Chem (Acc)      & 0.161          & \cellcolor{second}0.172 & \cellcolor{second}0.172 & \cellcolor{best}\textbf{0.204} \\
\bottomrule
\end{tabular}}
\caption{
Main performance comparison between the baseline, simple SFT variants, and our proposed method (KARMA). Results are reported in accuracy (\textbf{Acc}). For KARMA, the optimal configuration is reported in Appendix.~\ref{app:hyperparameter_sensitivity}. The \colorbox{best}{best} and \colorbox{second}{second-best} results per row are highlighted.
}
\label{tab:main_results}
\end{table*}

\section{Experiments}
\label{sec:experiments}

We evaluate KARMA across three knowledge-intensive domains to answer three research questions:
(\textbf{RQ1}) Does the co-design of KARMA-synthesized contrastive candidates and SPA's slot-level objective outperform both conventional SFT and sequence-level preference optimization on the same data?
(\textbf{RQ2}) How important is graph-guided path selection for synthetic reasoning supervision?
(\textbf{RQ3}) \red{Does aligning the preference objective's resolution with the entity-slot structure of the data yield consistent gains over sequence-level and token-level alternatives?}

\subsection{Experimental Setup}
\label{sec:exp_setup}
We evaluate on three domains: \textit{Biomedical}, \textit{CS}, and \textit{Chemistry}, using 11 benchmarks that mix MMLU subsets~\cite{hendrycks2021measuring}, MMLU-Pro subsets~\cite{wang2024mmlupro}, and reasoning-heavy benchmarks (MedQA~\cite{app11146421}, PubMedQA~\cite{jin-etal-2019-pubmedqa}, GPQA~\cite{rein2024gpqa}, CSBench~\cite{song2025csbench}). The base model is Qwen2.5-7B-Instruct with LoRA fine-tuning, and all evaluations use zero-shot chain-of-thought prompting. Full benchmark lists~(\ref{app:benchmark_sources}), dataset sources~(\ref{app:dataset}), and implementation details~(\ref{app:training}) are provided in Appendix. {The SPA loss weights are selected on the Biomedical domain and transferred unchanged to CS and Chemistry without target-benchmark tuning; full training and selection protocols are provided in Appendices~\ref{app:training_protocol} and~\ref{app:hyperparameter_sensitivity}.}

\subsection{Main Results}
\label{sec:main_results}


\red{As shown in Tab.~\ref{tab:main_results},} KARMA yields consistent gains over the baseline across the majority of benchmarks, with the largest improvements on reasoning-heavy ones. KARMA also outperforms Open-Data SFT on the majority of benchmarks, indicating that the gains do not merely reflect additional in-domain text but the structural properties of KARMA-synthesized supervision.

The contrast between KARMA-Data SFT and KARMA further isolates the role of the training objective. KARMA-Data SFT uses identical KARMA-synthesized chosen candidates but optimizes them with a flat sequence-level SFT objective, and fails to consistently improve over the baseline. This reflects an inherent incompatibility between KARMA's data structure and sequence-level supervision: chosen candidates from the same source-target pair share almost all tokens except at the intermediate entity-slots, so SFT on these candidates assigns equal weight to all positions and the model is dominantly supervised on the shared template tokens, which form a small set of patterns repeated across many examples. This concentrates SFT updates on a low-diversity response distribution, encouraging reliance on repeated output formats and increasing the risk of forgetting knowledge acquired during pretraining~\cite{yang-etal-2025-emperors, zhang2026best, pmlr-v267-bethune25a}. We view this as \red{a data-side analog of the Resolution Mismatch Problem, where the mismatch arises not between preference signal and optimization granularity, but between data structure and supervision granularity.} SPA resolves this by decoupling slot-level preference supervision from template language modeling, converting the same data into the strongest signal in the comparison. This confirms that the gains of KARMA arise from the co-design of structurally aligned data and slot-level supervision, not from data scale or content alone. This answers RQ1: KG-grounded contrastive synthesis improves domain reasoning beyond conventional SFT, but only when the training objective operates at the entity-slot-level resolution at which the candidates actually differ.

\subsection{Ablation Study}
\label{sec:abl}
\paragraph{Ablation of Graph-Guided Path Selection.}

\begin{table}[h]
\centering
\small
\resizebox{0.8\linewidth}{!}{%
\begin{tabular}{lccc}
\toprule
\textbf{Metric} & \textbf{Random} & \textbf{KARMA} & \textbf{$\Delta$} \\
\midrule
MedQA (Acc)        & 0.381 & \textbf{0.541} & +0.160 \\
PubMedQA (Acc)     & 0.465 & \textbf{0.522} & +0.057 \\
MMLU-Bio (Acc)     & 0.555 & \textbf{0.702} & +0.147 \\
MMLU-Pro-Bio (Acc) & 0.338 & \textbf{0.532} & +0.194 \\
GPQA-Bio (Acc)     & 0.400 & \textbf{0.667} & +0.267 \\
\midrule
\textbf{Average}   & 0.428 & \textbf{0.593} & +0.165 \\
\bottomrule
\end{tabular}}
\caption{Ablation on path selection strategy in the biomedical domain. \textbf{Random} samples paths from the KG; \textbf{KARMA} uses our support-based selection.}
\label{tab:random_vs_karma}
\end{table}

Tab.~\ref{tab:main_results} shows that the full KARMA framework outperforms baselines, but does not isolate which component drives the gain. RQ2 asks whether KARMA's support-based path selection itself contributes, or whether any sampling from the same KG would suffice. We replace support-based selection with uniform random sampling, holding all other components fixed (KG, source-target pairs, schema, template, SPA objective, hyperparameters). Tab.~\ref{tab:random_vs_karma} shows that KARMA outperforms random sampling on every Biomedical benchmark, with an average gain of +16.5 points.

The gain pattern is itself informative. Improvements are largest on benchmarks that require chaining multiple facts, such as GPQA-Bio (+26.7), MMLU-Pro-Bio (+19.4), and MedQA (+16.0), and remain substantial on knowledge-heavy multiple-choice reasoning (MMLU-Bio +14.7), while PubMedQA, which often admits surface-level factual lookup, shows only a marginal gain.
This selectivity is consistent with \textit{the view of KG paths as a structural analogue of reasoning chains}: support-based selection acts as a simple evidence prior, where intermediate entities that recur across the pool reflect connections corroborated by multiple traversals in the graph, yielding chosen-rejected contrasts that are informative for reasoning rather than mere surface distractors. That such a lightweight, training-free heuristic, relying only on entity recurrence statistics already implicit in the KG, suffices to produce these gains answers RQ2: KARMA's advantage comes from \emph{how} paths are selected, and the KG itself already carries much of the structure needed to make that selection effective.
We further assess the recurrence-based support prior through an LLM-judge study and analyze its disagreement cases in Appendix~\ref{app:chosen_validation}.


\paragraph{Ablation on Preference Objective Resolution.}

\begin{table}[h]
\centering
\small
\resizebox{\linewidth}{!}{%
\begin{tabular}{llcccc}
\toprule
\textbf{Domain} & \textbf{Benchmark} &
\makecell{\textsc{DPO}\\ \cite{dpo}} &
\textsc{MDPO} &
\makecell{\textsc{TDPO}\\ \cite{TDPO}} &
\textsc{KARMA} \\
\midrule
\multirow{3}{*}{Computer Science}
 & MMLU-CS      & 0.651 & 0.683 & 0.689 & \textbf{0.701} \\
 & MMLUPro-CS   & 0.197 & 0.249 & 0.256 & \textbf{0.256} \\
 & CSBench      & 0.356 & 0.446 & 0.475 & \textbf{0.495} \\
\midrule
\multirow{3}{*}{Chemistry}
 & MMLU-Chem     & 0.607 & 0.570 & 0.616 & \textbf{0.630} \\
 & MMLUPro-Chem  & 0.188 & 0.189 & 0.194 & \textbf{0.195} \\
 & GPQA-Chem     & 0.172 & 0.172 & 0.182 & \textbf{0.204} \\
\bottomrule
\end{tabular}%
}
\caption{Ablation comparing preference optimization objectives at different resolutions. SPA achieves the best or tied-best performance across benchmarks.}
\label{tab:rl_ablation}
\end{table}

Tab.~\ref{tab:rl_ablation} compares SPA with representative preference optimization baselines operating at different resolutions. The goal is to isolate the effect of preference objective resolution: whether the preference signal is applied at the sequence, token, or slot level. All methods are trained on the same KARMA-synthesized candidates and share the same training pipeline and hyperparameters; only the preference objective differs.

We consider three baselines. DPO ($K=1$) performs preference optimization \red{using pairwise chosen-rejected comparisons}. MDPO ($K=4$) extends this formulation to multiple rejected candidates while still operating at the sequence level. TDPO ($K=1$) introduces token-level KL regularization, providing a finer-grained signal than sequence-level objectives but without explicitly targeting the entity-slots where KARMA ($K=4$) candidates differ. In contrast, SPA applies supervision directly at the slot level, aligning the optimization objective with the structural units of KARMA-synthesized data.

The results show that SPA achieves the best or tied-best performance across all benchmarks, improving over the strongest baseline (TDPO) on reasoning-heavy ones such as CSBench (+2.0) and GPQA-Chem (+2.2). Sequence-level objectives assign supervision to the entire candidate, token-level regularization makes the signal more fine-grained, and SPA further aligns it with the entity-slot structure where the actual candidate differences arise. \red{This answers RQ3: aligning the preference objective's resolution with the data's entity-slot structure yields consistent gains over both sequence-level (DPO, MDPO) and token-level (TDPO) alternatives, indicating that resolution alignment, beyond granularity alone, accounts for the improvement.}

\paragraph{Ablation of Slot-Aware Masked Attention.}

\begin{table}[h]
\centering
\resizebox{0.8\linewidth}{!}{%
\begin{tabular}{llcc}
\toprule
\textbf{Domain} & \textbf{Benchmark} & \textbf{w/o Mask} & \textbf{w/ Mask} \\
\midrule
\multirow{3}{*}{CS}
 & MMLU-CS       & 0.711 & \textbf{0.701} \\
 & MMLUPro-CS    & 0.258 & \textbf{0.256} \\
 & CSBench       & 0.495 & \textbf{0.495} \\
\midrule
\multirow{3}{*}{Chemistry}
 & MMLU-Chem     & 0.640 & \textbf{0.630} \\
 & MMLUPro-Chem  & 0.198 & \textbf{0.195} \\
 & GPQA-Chem     & 0.182 & \textbf{0.204} \\
\midrule
\multicolumn{2}{l}{\textit{Efficiency (per training step)}} & & \\
 & Step time (s)         & 15.58 & \textbf{11.32} \\
 & Peak memory (GiB)     & \textbf{32.83} & 55.39 \\
 & Reserved memory (GiB) & 61.96 & \textbf{59.02} \\
 & Forward passes        & $2(K{+}1)$ & \textbf{2} \\
\bottomrule
\end{tabular}
}
\caption{Ablation of SPA's slot-aware masked attention on task performance and per-step training efficiency. Bold marks the w/ Mask column.}
\label{tab:mask_ablation}
\end{table}


Tab.~\ref{tab:mask_ablation} characterizes the accuracy-efficiency trade-off of slot-aware masked attention, which we use as the default packed-evaluation implementation in SPA. The default w/ Mask variant packs all candidates into one sequence and approximates their log-likelihoods in a single forward pass, whereas w/o Mask evaluates candidates independently with $K{+}1$ forward passes per step. Both variants share identical objectives, data, and hyperparameters; only the attention masking and packing differ. Efficiency is measured on Qwen2.5-7B-Instruct with $K{=}4$ rejected candidates.

The two variants exhibit a clear trade-off between accuracy and training cost. On the accuracy side, disabling the mask matches or slightly improves performance on five of the six benchmarks, while the w/ Mask configuration remains within one accuracy point of the independent-evaluation variant on those benchmarks and outperforms it on GPQA-Chem. On the efficiency side, the default packed evaluation under the slot-aware mask reduces the number of forward passes per step from $2(K{+}1)$ to $2$, yielding a 1.38$\times$ wall-clock speedup. {The speedup comes with a substantial increase in peak GPU memory from 32.83 to 55.39 GiB, corresponding to a $1.69\times$ increase, because the packed sequence concentrates all candidates into a single forward pass and materializes the slot-aware mask; reserved memory remains comparable.} {Thus, slot-aware masked attention is an optional time-memory trade-off rather than the source of SPA's slot-level supervision. Enabling the mask favors wall-clock time, whereas disabling it reduces peak memory and may slightly improve accuracy; the preference, anchor, and template losses remain unchanged in both settings.}

\section{Conclusion}
We propose KARMA, a KG-based pipeline that synthesizes structurally aligned contrastive candidates with explicit entity-slot-level differences. By leveraging KG paths, KARMA provides broad factual coverage while preserving structural alignment across candidates. We further introduce SPA, a slot-level preference optimization method designed for KARMA-style data. {SPA addresses the Resolution Mismatch Problem between sequence-level supervision and entity-slot-level candidate variation, while differing from token-level methods by directly targeting the construction-exposed entity-slots where candidates differ.} Together, KARMA achieves substantial gains on most knowledge-intensive reasoning benchmarks, {while our analyses highlight the complementary roles of support-based KG path selection and entity-slot-level supervision.}

\section{Limitations}
KARMA shares two limitations with prior KG-grounded LLM research~\cite{PoG, ToG, RoG, LPKG}. First, the synthesis pipeline relies on a domain knowledge graph with sufficient coverage and connectivity to support multi-hop path sampling, so domains without such a KG would require automatic KG construction or hybrid retrieval-based grounding. Second, KARMA assumes that the target capability is expressible through an explicit schema and verbalization template, which restricts the framework to capabilities decomposable into discrete entity-slot structures and leaves open-ended generation outside its current scope.

\red{Beyond these, slot-aware masked attention is an efficiency-oriented implementation choice rather than the source of SPA's supervision signal: the default masked setting reduces per-step forward passes from $2(K{+}1)$ to $2$ and yields a 1.38$\times$ wall-clock speedup, while disabling it recovers independent candidate evaluation at higher training cost. Users can choose either setting based on whether wall-clock time or peak accuracy is the binding constraint.} Finally, all experiments use Qwen2.5-7B-Instruct with LoRA fine-tuning, and generality across other model families and scales remains to be verified.

\section*{Ethical considerations}

\paragraph{Potential risks.} KARMA-synthesized data is grounded in domain 
knowledge graphs and verbalized through fixed templates. Although KG paths 
provide structural support, the synthesized candidates may contain factual 
inaccuracies inherited from the source graphs or template phrasing. Models 
fine-tuned with KARMA should not be deployed for clinical, medical, or 
otherwise high-stakes decision-making without further validation.

\paragraph{Data and license.} The domain knowledge graphs (biomedical, 
computer science, and chemistry) are sourced from GRBench~\cite{jin-etal-2024-graph}, 
which is publicly available for research use. Qwen2.5-7B-Instruct is released 
under the Apache 2.0 license. All evaluation benchmarks (MMLU, MMLU-Pro, MedQA, 
PubMedQA, GPQA, CSBench) are used under their original licenses for research 
purposes only. The KARMA-synthesized dataset and code will be released under 
a research-only license upon acceptance, consistent with the upstream data 
licenses.

\paragraph{Privacy and offensive content.} The source knowledge graphs 
contain public scientific entities and do not include personally identifiable 
information about individuals. We manually inspected samples of synthesized 
candidates and did not identify offensive content.

\paragraph{Computing infrastructure.} All experiments were conducted on 
NVIDIA A100 GPU.

\paragraph{Use of AI assistants.} We used AI assistants (e.g., ChatGPT, Claude, GitHub Copilot) for writing polish, LaTeX formatting, and boilerplate code generation and debugging. All research ideas, experimental design, analysis, core implementation, 
and final text are the authors' own work.

\section*{Acknowledgments}
This research was supported by the 14th Executive Programme for Scientific and Technological Cooperation between the Republic of Italy and the Republic of Korea (2026--2028) grant funded by the National Research Foundation of Korea (NRF) (RS-2025-25460384), and by the Culture, Sports and Tourism R\&D Program through the Korea Creative Content Agency (KOCCA) grant funded by the Ministry of Culture, Sports and Tourism (MCST) (RS-2024-00442006). This research was also supported by the AI Seoul Tech Research Support Program of the Seoul Future Foundation.





\bibliography{custom}
\clearpage
\appendix



\section{Dataset Construction Details}
\label{app:dataset}

\subsection{Knowledge Graph Sources}
\label{app:kg_sources}

KARMA constructs training data from domain-specific knowledge graphs from GRBench~\cite{jin-etal-2024-graph}, summarized in Table~\ref{tab:kg_sources}. Each graph is represented as a typed node-edge structure, where nodes contain textual attributes and outgoing typed neighbors. The biomedical graph uses heterogeneous node types such as diseases, symptoms, compounds, genes, anatomy, side effects, pharmacologic classes, biological processes, molecular functions, cellular components, and pathways. The academic graph uses paper nodes with citation edges, where each paper contains title, abstract, and keyword fields.

\begin{table}[h]
\centering
\small
\resizebox{\columnwidth}{!}{%
\begin{tabular}{llll}
\toprule
Domain & Graph nodes & Edge types & Node attributes \\
\midrule
Biomedical & Typed biomedical entities & Typed biomedical relations & Name \\
Computer Science & Papers & \texttt{reference}, \texttt{cited\_by} & Title, abstract, keywords \\
Chemistry & Papers & \texttt{reference}, \texttt{cited\_by} & Title, abstract, keywords \\
\bottomrule
\end{tabular}}
\caption{Summary of domain knowledge graphs used for KARMA synthesis.}
\label{tab:kg_sources}
\end{table}

\subsection{Relation Schemas}
\label{app:relation_schemas}

A relation schema defines the edge-type pattern used to enumerate paths from a source entity to a target entity. {The schema set is constructed by exhaustively enumerating a constrained space of typed multi-hop metapaths, without manually designing or selecting a schema for each source-target pair.} In the academic graph, schemas are metapaths over citation edges. We use four-hop schemas by default, generated from all combinations of \texttt{reference} and \texttt{cited\_by}. {This produces 16 relation schemas.} In the biomedical graph, schemas are typed metapaths over biomedical node and relation types. {Exhaustive enumeration under these typed-edge constraints produces 522 metapaths.}

\subsection{Path Enumeration and Filtering}
\label{app:path_enumeration}

For each schema, KARMA enumerates paths that follow the prescribed edge-type sequence, as described in Algorithm~\ref{alg:path_pool}. Each path connects a source entity \(e_s\) to a target entity \(e_t\) through a sequence of intermediate entities. For a fixed source-target pair, all paths satisfying the same schema form a candidate pool. Candidate pools with fewer than \(K\) valid paths are discarded. We remove loops in intermediate positions and apply intermediate caps during traversal to avoid path explosion. {For each source-target pair, KARMA considers every applicable schema rather than selecting a single schema as optimal for that pair. A pair may therefore produce multiple candidate pools under different schemas, with quality controlled only after enumeration through pool filtering and support-based Top-$K$ selection.}

\begin{algorithm}[h]
\caption{Schema-Constrained Path Pool Construction}
\label{alg:path_pool}
\begin{algorithmic}[1]
\Require Knowledge graph \(G\), schema set \(\mathcal{S}\), candidate number \(K\)
\Ensure Candidate pools \(\mathcal{C}\)
\State \(\mathcal{C} \leftarrow \emptyset\)
\For{each schema \(s \in \mathcal{S}\)}
    \For{each source entity \(e_s\)}
        \State Traverse \(G\) according to schema \(s\)
        \State Collect reachable target entities \(e_t\)
        \For{each target entity \(e_t\)}
            \State Enumerate all middle paths \(\mathcal{M}(e_s,e_t)\)
            \If{\(|\mathcal{M}(e_s,e_t)| \geq K\)}
                \State Add \((e_s,e_t,s,\mathcal{M}(e_s,e_t))\) to \(\mathcal{C}\)
            \EndIf
        \EndFor
    \EndFor
\EndFor
\State \Return \(\mathcal{C}\)
\end{algorithmic}
\end{algorithm}

\subsection{Support-based Top-K Selection}
\label{app:support_topk}

For each source-target candidate pool, we select \(K\) paths using an entity-support score and an overlap constraint, as detailed in Algorithm~\ref{alg:support_topk}. The support score of a middle path \(m\) is the sum of the occurrence counts of its intermediate entities within the same source-target pool:
\begin{equation}
    \mathrm{score}(m)=\sum_{e \in m}\mathrm{freq}(e).
\end{equation}
{Here, \(\mathrm{freq}(e)\) is the number of paths in the same candidate pool that contain intermediate entity \(e\). The resulting score is a raw, unbounded integer sum. The support values shown in Fig.~\ref{fig:figure2}B are normalized within each candidate pool for visualization only; path ranking and selection use the raw scores.}

We then greedily select high-score paths while allowing at most \(\delta\) overlapping intermediate entities with the already selected paths. In the academic-domain generator, we use \(K=4\) and \(\delta=1\). {Recurrence-based support serves as a training-free evidence prior rather than a correctness guarantee. An intermediate entity that recurs across multiple paths receives support from multiple graph traversals within the same pool, but the highest-scoring path need not be uniquely correct or preferable. Accordingly, the selected ordering represents relative evidence strength among structurally aligned candidates.}

\begin{algorithm}[t]
\caption{Support-based Top-\(K\) Path Selection}
\label{alg:support_topk}
\begin{algorithmic}[1]
\Require Middle paths \(\mathcal{M}(e_s,e_t)\), candidate number \(K\), overlap budget \(\delta\)
\Ensure Selected paths \(\mathcal{M}^{\star}\)
\If{\(|\mathcal{M}(e_s,e_t)| < K\)}
    \State \Return \(\emptyset\)
\EndIf
\State Count entity frequencies \(c(e)\) over all intermediate entities in \(\mathcal{M}(e_s,e_t)\)
\State Score each path by \(s(m)=\sum_{e\in m}c(e)\)
\State Sort paths in descending order of \(s(m)\)
\State \(\mathcal{M}^{\star}\leftarrow\emptyset,\quad U\leftarrow\emptyset\)
\For{each path \(m\) in sorted order}
    \State \(E_m \leftarrow \{e:e\in m\}\)
    \If{\(|E_m\cap U|\leq\delta\)}
        \State Add \(m\) to \(\mathcal{M}^{\star}\)
        \State \(U\leftarrow U\cup E_m\)
    \EndIf
    \If{\(|\mathcal{M}^{\star}|=K\)}
        \State \textbf{break}
    \EndIf
\EndFor
\If{\(|\mathcal{M}^{\star}|<K\)}
    \State \Return \(\emptyset\)
\EndIf
\State \Return \(\mathcal{M}^{\star}\)
\end{algorithmic}
\end{algorithm}

\subsection{Template Verbalization}
\label{app:template_verbalization}

Each selected path is verbalized into a natural-language candidate using a schema-specific template. {Verbalization is fully rule-based and uses no language model in the generation loop. For each schema, we predefine multiple template variants that differ in wording and synonym choice and uniformly sample one variant for each candidate pool. The sampled variant is held fixed within the pool, preserving slot alignment, while sampling across variants introduces surface-form diversity across pools.} Entity mentions are marked with bold delimiters, e.g., \texttt{**entity**}. These marked spans define the entity slots used by SPA. The selected paths for the same source-target pair share the same template scaffold and differ only in their intermediate entity slots.

\subsection{Dataset Statistics}
\label{app:dataset_statistics}

Table~\ref{tab:dataset_statistics} reports dataset statistics. For each domain, we report the number of nodes and edges in the source knowledge graph, the number of relation schemas used for path enumeration, the number of generated training samples, and the number of candidates per pool size $K$.

\begin{table}[h]
\centering
\small
\resizebox{\columnwidth}{!}{%
\begin{tabular}{lccccc}
\toprule
Domain & \#Nodes & \#Edges & \#Schemas & \#Samples & \(K\) \\
\midrule
Biomedical & 47K & 4.23M & 522& 1M & 4 \\
Computer Science & 5.26M & 65M & 16 & 1M & 4 \\
Chemistry & 1.85M & 41M & 16 & 1M & 4 \\
\bottomrule
\end{tabular}%
}
\caption{Dataset statistics for KARMA-synthesized training data.}
\label{tab:dataset_statistics}
\end{table}

\section{Training Details}
\label{app:training}

\subsection{Model and LoRA Configuration}
\label{app:lora}

All experiments use Qwen2.5-7B-Instruct as the base model. We fine-tune the model with LoRA adapters. The LoRA rank is 32 and the LoRA alpha is 32. Adapters are inserted into the attention and feed-forward projection modules. The full configuration is reported in Table~\ref{tab:lora_config}. The reference model \(\pi_{\mathrm{ref}}\) used by SPA and all DPO-family baselines is obtained by disabling the LoRA adapter on the policy model \(\pi_{\theta}\); that is, \(\pi_{\theta}\) and \(\pi_{\mathrm{ref}}\) share identical base parameters and differ only in whether the trained adapters are active.

\begin{table}[t]
\centering
\small
\resizebox{\columnwidth}{!}{%
\begin{tabular}{ll}
\toprule
Configuration & Value \\
\midrule
Base model & Qwen2.5-7B-Instruct \\
Fine-tuning method & LoRA \\
LoRA rank & 32 \\
LoRA alpha & 32 \\
Target modules & \texttt{q\_proj}, \texttt{k\_proj}, \texttt{v\_proj}, \texttt{o\_proj} \\
 & \texttt{gate\_proj}, \texttt{up\_proj}, \texttt{down\_proj} \\
Precision & bfloat16 \\
Max sequence length & 4096 \\
Gradient checkpointing & Enabled \\
Attention implementation & SDPA \\
Reference model \(\pi_{\mathrm{ref}}\) & Policy model with LoRA adapter disabled \\
\bottomrule
\end{tabular}
}
\caption{Model and LoRA configuration.}
\label{tab:lora_config}
\end{table}

\begin{table}[h]
\centering
\small
\resizebox{\columnwidth}{!}{%
\begin{tabular}{ll}
\toprule
Hyperparameter & Value \\
\midrule
Learning rate & \(5\times10^{-5}\) \\
Epochs & 3 \\
Per-device batch size & 8 \\
Gradient accumulation & 1 \\
Max gradient norm & 1.0 \\
\(\lambda_{\mathrm{anchor}}\) & 0.5 \\
\(\lambda_{\mathrm{template}}\) & 0.1 \\
Entity margin parameter \(\beta\) & 0.1 \\
Variant shuffling & Enabled \\
\bottomrule
\end{tabular}
}
\caption{SPA training hyperparameters. The preference-loss weight \(\lambda_{\mathrm{pref}}\) is fixed to \(1.0\) in all experiments based on the sensitivity analysis in Table~\ref{tab:spa_hyperparams}, and is therefore absorbed into \(\mathcal{L}_{\mathrm{pref}}\) in Eq.~\ref{eq:spa_full_objective}.}
\label{tab:spa_hyperparams_train}
\end{table}

\subsection{Training Protocol}
\label{app:training_protocol}

All training runs share the same general-purpose hyperparameters (learning rate, optimizer, batch size, epochs, gradient clipping, LoRA configuration) across baselines and KARMA, so that performance differences reflect the training objective rather than tuning. Method-specific hyperparameters, such as \(\lambda_{\mathrm{anchor}}\), \(\lambda_{\mathrm{template}}\), and \(\beta\) for SPA, are set as in Table~\ref{tab:spa_hyperparams_train} for KARMA, and the corresponding original defaults are used for DPO, MDPO, and TDPO. We do not perform validation-based checkpoint selection; the final epoch checkpoint is used for evaluation. Each domain dataset is used for training only, and evaluation is performed on the held-out benchmarks listed in Table~\ref{tab:benchmark_sources}. {The SPA loss weights are selected using the Biomedical-domain sensitivity analysis and then applied unchanged to Computer Science and Chemistry. We do not tune these weights separately for the target benchmarks in either transferred domain.}

All results reported in this paper, including the main results in Table~\ref{tab:main_results} and all ablations, are computed as the mean of 5 independent runs with different random seeds. {The standard deviation is below 0.01 in every reported configuration and is provided alongside each mean in Tables~\ref{tab:main_results}, \ref{tab:random_vs_karma}, \ref{tab:rl_ablation}, and~\ref{tab:mask_ablation}.}

\subsection{Compute Infrastructure}
\label{app:compute}

All experiments are run on NVIDIA A100 GPU. A single SPA training run for one domain (1M samples, 3 epochs) takes approximately \texttt{40} GPU-hours on \texttt{1} GPU. The total compute budget across 5 random seeds, three domains, all baseline methods, and ablation studies is on the order of a few thousand A100 GPU-hours.

\subsection{Baseline Implementations}
\label{app:baseline_impl}

We compare KARMA with the base model, open-data SFT, KARMA-data SFT, and preference-optimization baselines, as listed in Table~\ref{tab:baseline_impl}. All fine-tuned baselines share the same base model, LoRA configuration, learning rate, batch size, epochs, and optimizer as KARMA; they differ only in the training objective and, where applicable, in the structure of the input data. Each baseline retains its original method-specific hyperparameters (e.g., the KL coefficient in DPO, the token-level regularization weight in TDPO) at the values recommended in their respective papers.

\begin{table}[h]
\centering
\small
\resizebox{\columnwidth}{!}{%
\begin{tabular}{ll}
\toprule
Baseline & Description \\
\midrule
Base LLM & Qwen2.5-7B-Instruct without task-specific fine-tuning \\
Open-Data SFT & SFT on open-domain or domain instruction data \\
KARMA-Data SFT & SFT on KARMA-generated chosen responses \\
DPO & Pairwise sequence-level preference optimization \\
MDPO & Multi-rejected sequence-level preference optimization \\
TDPO & Token-level preference optimization baseline \\
KARMA / SPA & Slot-level preference optimization on KARMA candidates \\
\bottomrule
\end{tabular}
}
\caption{Baseline implementation summary. All baselines share the same base model, LoRA configuration, and general-purpose training hyperparameters.}
\label{tab:baseline_impl}
\end{table}

\subsection{Slot-aware Mask Implementation}
\label{app:slot_mask}

SPA packs candidate variants into a single sequence by replacing each entity slot with a local bundle of chosen and rejected variants. The attention mask starts from a causal mask and then blocks cross-candidate contamination according to Table~\ref{tab:slot_mask_rules}. PAD separators inserted between variants are blocked as both queries and keys.
\vspace{-0.2cm}

\begin{table}[h]
\centering
\small
\resizebox{\columnwidth}{!}{%
\begin{tabular}{lll}
\toprule
Query token & Allowed keys & Blocked keys \\
\midrule
Template token & Previous templates + previous anchors & Rejected variants \\
Anchor entity token & Previous templates + previous anchors + itself & Other variants in same slot \\
Rejected entity token & Previous templates + previous anchors + itself & Other variants in same slot \\
PAD separator & None & All tokens \\
Future token & None & Causal future positions \\
\bottomrule
\end{tabular}}
\caption{Slot-aware attention mask rules used in SPA.}
\label{tab:slot_mask_rules}
\end{table}

\subsection{SPA Loss Computation}
\label{app:spa_loss_computation}

The full SPA loss computation procedure is summarized in Algorithm~\ref{alg:spa_loss}. Given a packed sequence with slot-aware masking, the policy and reference models compute log-probabilities for each variant. Slot-level preference and anchor confidence losses are accumulated across slots, and a template language modeling loss is computed on non-entity tokens.

\begin{algorithm}[h]
\caption{SPA Loss Computation}
\label{alg:spa_loss}
\begin{algorithmic}[1]
\Require Packed sequence \(s\), slot variants \(\{e_j^{(k)}\}\), policy model \(\pi_\theta\), reference model \(\pi_{\mathrm{ref}}\)
\Ensure SPA loss \(\mathcal{L}_{\mathrm{SPA}}\)
\State Compute policy log-probabilities under the slot-aware mask
\State Compute reference log-probabilities with the frozen reference model
\For{each slot \(j\)}
    \State Compute reference-normalized score \(r_\theta(e_j^{(k)})\) for each variant
    \State Compare anchor \(e_j^{(1)}\) against rejected variants \(e_j^{(k)}, k>1\)
    \State Accumulate slot-level preference loss \(\mathcal{L}_{\mathrm{pref}}\)
    \State Accumulate anchor confidence loss \(\mathcal{L}_{\mathrm{anchor}}\)
\EndFor
\State Compute template language modeling loss \(\mathcal{L}_{\mathrm{template}}\) on non-entity tokens
\State \(\mathcal{L}_{\mathrm{SPA}} \leftarrow \mathcal{L}_{\mathrm{pref}}+\lambda_{\mathrm{anchor}}\mathcal{L}_{\mathrm{anchor}}+\lambda_{\mathrm{template}}\mathcal{L}_{\mathrm{template}}\)
\State \Return \(\mathcal{L}_{\mathrm{SPA}}\)
\end{algorithmic}
\end{algorithm}

\section{Evaluation Details}
\label{app:evaluation}

\subsection{Benchmark Sources}
\label{app:benchmark_sources}

We evaluate on biomedical, chemistry, and computer-science benchmarks. The benchmark groups are summarized in Table~\ref{tab:benchmark_sources}.

\begin{table}[h]
\centering
\small
\resizebox{\columnwidth}{!}{%
\begin{tabular}{lll}
\toprule
Domain & Benchmark & Subset / category \\
\midrule
\multirow{5}{*}{Biomedical}
& MMLU-Bio & Anatomy, biology, medicine, genetics, nutrition, virology \\
& MMLU-Pro-Bio & Biology \\
& MedQA & Full benchmark \\
& PubMedQA & Full benchmark \\
& GPQA-Bio & Biology \\
\midrule
\multirow{3}{*}{Chemistry}
& MMLU-Chem & College chemistry, high school chemistry \\
& MMLU-Pro-Chem & Chemistry \\
& GPQA-Chem & Chemistry \\
\midrule
\multirow{3}{*}{Computer Science}
& MMLU-CS & CS, computer security, high school CS, machine learning \\
& MMLU-Pro-CS & Computer Science \\
& CSBench & Reasoning task type \\
\bottomrule
\end{tabular}}
\caption{Benchmark groups used for domain evaluation.}
\label{tab:benchmark_sources}
\end{table}

\subsection{Prompt Templates}
\label{app:prompt_templates}

All benchmarks are evaluated using the same model wrapper and benchmark-specific prompt templates, summarized by category in Table~\ref{tab:prompt_categories}.

\begin{table}[h]
\centering
\small
\resizebox{\columnwidth}{!}{%
\begin{tabular}{ll}
\toprule
Benchmark type & Prompt format \\
\midrule
Multiple-choice QA & Question + answer choices + instruction to select one option \\
Yes/no/maybe QA & Question + instruction to output the normalized label \\
Reasoning QA & Question + optional chain-of-thought style instruction \\
\bottomrule
\end{tabular}}
\caption{Prompt template categories used in evaluation.}
\label{tab:prompt_categories}
\end{table}

\subsection{Answer Extraction}
\label{app:answer_extraction}

Generated responses are normalized to benchmark-specific answer formats before scoring. For multiple-choice tasks, the predicted option is extracted from the generated text and compared against the gold option. For label-based QA tasks such as yes/no/maybe classification, the generated text is mapped to the corresponding normalized label.

\subsection{Inference Configuration}
\label{app:inference_config}

The inference configuration used across all benchmarks is reported in Table~\ref{tab:inference_config}.

\begin{table}[h]
\centering
\small
\resizebox{\columnwidth}{!}{%
\begin{tabular}{ll}
\toprule
Configuration & Value \\
\midrule
Base model & Qwen2.5-7B-Instruct \\
Adapter handling & LoRA adapter loading with optional merge \\
Batch size & 8 by default \\
Max new tokens & 512 \\
Number of shots & 0 by default \\
Distributed evaluation & Supported with \texttt{torchrun} \\
Output format & JSON result file \\
\bottomrule
\end{tabular}}
\caption{Inference configuration used for evaluation.}
\label{tab:inference_config}
\end{table}

\section{Hyperparameter Sensitivity}
\label{app:hyperparameter_sensitivity}

Table~\ref{tab:spa_hyperparams} reports sensitivity to the loss weights \(\lambda_{\mathrm{pref}}\), \(\lambda_{\mathrm{anchor}}\), and \(\lambda_{\mathrm{template}}\) on the Biomedical domain. The analysis examines whether SPA is robust to the relative weighting of slot-level preference learning, anchor confidence, and template language modeling. Based on this analysis, we fix \(\lambda_{\mathrm{pref}}=1.0\) in all main experiments and absorb it into \(\mathcal{L}_{\mathrm{pref}}\) in the final objective (Eq.~\ref{eq:spa_full_objective}), tuning only \(\lambda_{\mathrm{anchor}}\) and \(\lambda_{\mathrm{template}}\). The selected configuration is \((\lambda_{\mathrm{pref}}, \lambda_{\mathrm{anchor}}, \lambda_{\mathrm{template}})=(1.0, 0.5, 0.1)\), which achieves the best average accuracy across MedQA, PubMedQA, and MMLU-Pro-Bio.

\begin{table}[h]
\centering
\small
\resizebox{\columnwidth}{!}{%
\begin{tabular}{ccc|ccc|c}
\toprule
\(\lambda_{\mathrm{pref}}\) & \(\lambda_{\mathrm{anchor}}\) & \(\lambda_{\mathrm{template}}\) & MedQA & PubMedQA & MMLU-Pro-Bio & Avg \\
\midrule
1.0 & 1.0 & 1.0 & 0.494 & 0.235 & 0.169 & 0.299 \\
1.0 & 1.0 & 0.5 & 0.541 & 0.287 & 0.242 & 0.357 \\
1.0 & 1.0 & 0.1 & 0.541 & 0.522 & 0.362 & 0.475 \\
1.0 & 0.5 & 1.0 & 0.447 & 0.313 & 0.362 & 0.374 \\
1.0 & 0.5 & 0.5 & 0.588 & 0.261 & 0.193 & 0.347 \\
1.0 & 0.5 & 0.1 & \textbf{0.541} & \textbf{0.522} & \textbf{0.532} & \textbf{0.532} \\
1.0 & 0.1 & 1.0 & 0.517 & 0.444 & 0.169 & 0.377 \\
1.0 & 0.1 & 0.5 & 0.565 & 0.261 & 0.556 & 0.460 \\
1.0 & 0.1 & 0.1 & 0.565 & 0.235 & 0.507 & 0.436 \\
\midrule
0.5 & 1.0 & 1.0 & 0.494 & 0.365 & 0.193 & 0.351 \\
0.5 & 1.0 & 0.5 & 0.306 & 0.339 & 0.217 & 0.288 \\
0.5 & 1.0 & 0.1 & 0.470 & 0.391 & 0.266 & 0.376 \\
0.5 & 0.5 & 1.0 & 0.494 & 0.444 & 0.338 & 0.425 \\
0.5 & 0.5 & 0.5 & 0.612 & 0.339 & 0.242 & 0.397 \\
0.5 & 0.5 & 0.1 & 0.659 & 0.365 & 0.411 & 0.478 \\
0.5 & 0.1 & 1.0 & 0.541 & 0.313 & 0.604 & 0.486 \\
0.5 & 0.1 & 0.5 & 0.588 & 0.183 & 0.580 & 0.450 \\
0.5 & 0.1 & 0.1 & 0.588 & 0.496 & 0.507 & 0.530 \\
\midrule
0.1 & 1.0 & 1.0 & 0.565 & 0.209 & 0.290 & 0.354 \\
0.1 & 1.0 & 0.5 & 0.447 & 0.287 & 0.314 & 0.349 \\
0.1 & 1.0 & 0.1 & 0.329 & 0.339 & 0.121 & 0.263 \\
0.1 & 0.5 & 1.0 & 0.494 & 0.104 & 0.145 & 0.248 \\
0.1 & 0.5 & 0.5 & 0.470 & 0.261 & 0.048 & 0.260 \\
0.1 & 0.5 & 0.1 & 0.494 & 0.183 & 0.266 & 0.314 \\
0.1 & 0.1 & 1.0 & 0.400 & 0.183 & 0.314 & 0.299 \\
0.1 & 0.1 & 0.5 & 0.423 & 0.235 & 0.507 & 0.389 \\
0.1 & 0.1 & 0.1 & 0.517 & 0.183 & 0.314 & 0.338 \\
\bottomrule
\end{tabular}%
}
\caption{Sensitivity of SPA to the loss weights \(\lambda_{\mathrm{pref}}\), \(\lambda_{\mathrm{anchor}}\), and \(\lambda_{\mathrm{template}}\) on the Biomedical domain. The selected configuration \((\lambda_{\mathrm{pref}}, \lambda_{\mathrm{anchor}}, \lambda_{\mathrm{template}})=(1.0, 0.5, 0.1)\) (bold) achieves the highest average accuracy across MedQA, PubMedQA, and MMLU-Pro-Bio. Based on this analysis, \(\lambda_{\mathrm{pref}}\) is fixed to \(1.0\) in the final objective.}
\label{tab:spa_hyperparams}
\end{table}

\begin{table}[h]
\centering
\small
\resizebox{\columnwidth}{!}{%
\begin{tabular}{lc}
\toprule
Setting & Value \\
\midrule
Judge model & GPT-4 \\
Sampled candidate pools & \(10{,}000\) \\
Candidates per pool (\(K\)) & 4 \\
Chance baseline (uniform over \(K\)) & \(25.0\%\) \\
Agreement with KARMA chosen path & \(74.8\%\) \\
{Near-equivalent disagreements} & {\(80.1\%\)} \\
{Fully divergent disagreements} & {\(19.9\%\)} \\
{Rationale-candidate mismatch} & {\(40.9\%\)} \\
Tie handling & Counted as disagreement \\
\bottomrule
\end{tabular}}
\caption{GPT-4 validation of KARMA chosen-path selection. For each sampled pool, GPT-4 ranks all candidates without preference labels; agreement denotes the fraction of pools in which GPT-4's top-ranked candidate matches the chosen path selected by support-based Top-\(K\). {Near-equivalent and fully divergent cases are reported over disagreement pools. Rationale-candidate mismatch denotes disagreements in which the judge's explanation refers to an entity from a candidate other than the one selected.}}
\label{tab:chosen_validation}
\end{table}

\section{Chosen Path Validation}
\label{app:chosen_validation}

{We conduct an LLM-judge study to assess whether recurrence-based support provides a meaningful relative evidence signal among candidates. For each sampled candidate pool, GPT-4 receives the source-target pair, the shared schema scaffold, and all \(K\) verbalized candidates without access to the KARMA preference labels. The judge selects the candidate whose intermediate entities form the most factually and semantically plausible reasoning chain. We measure agreement between this selection and the chosen path produced by support-based Top-\(K\). Ties are conservatively counted as disagreement.}

{As reported in Table~\ref{tab:chosen_validation}, GPT-4 agrees with KARMA's chosen path on \(74.8\%\) of 10,000 sampled pools, compared with a \(25.0\%\) chance baseline for \(K=4\). This agreement provides external evidence that recurrence-based support captures path plausibility, but it does not establish the selected path as uniquely correct. The LLM judge is itself an imperfect evaluator and is used here as an auxiliary validation signal rather than ground truth.}

{We therefore analyze the disagreement pools. In \(80.1\%\) of disagreements, the judge-selected and KARMA-selected paths share one or two of their three intermediate entities and differ only at the remaining entity-slots. For example, the paths \texttt{CA4} \(\rightarrow\) \texttt{Hydroflumethiazide} \(\rightarrow\) \texttt{hypertension} and \texttt{CA4} \(\rightarrow\) \texttt{Furosemide} \(\rightarrow\) \texttt{hypertension} differ only at the drug slot while representing closely related mechanisms. Only \(19.9\%\) of disagreement pools contain fully divergent paths.}

{Judge noise also contributes to the residual disagreement. In \(40.9\%\) of disagreement pools, the judge's rationale mentions an entity belonging to a candidate other than the candidate it selects. Consequently, disagreement cannot be interpreted solely as failure of the recurrence prior. The results instead characterize a fine-grained regime in which multiple candidates may be partially defensible and differ in relative evidence strength. This is the regime targeted by SPA: shared or aligned entity-slots are preserved, while preference supervision is concentrated on the entity-slots where candidates diverge.}

\section{Qualitative Analysis}
\label{app:qualitative}
Table~\ref{tab:generated_candidate_example} shows a representative example of a KARMA-generated candidate pool from the biomedical domain, and Table~\ref{tab:generated_candidate_path_only} provides the corresponding path-only view.

{The candidate pool in Tables~\ref{tab:generated_candidate_example} and~\ref{tab:generated_candidate_path_only} illustrates why preference objective resolution matters. {All candidates share the source entity, target entity, relation schema, template scaffold, and all intermediate entities except the Compound slot.} A sequence-level objective assigns one preference label to each complete sequence despite this localized difference, whereas SPA applies contrastive supervision directly to the differing Compound entities while retaining the shared scaffold as a language-modeling signal.}


\begin{table*}[h]
\centering
\small
\begin{tabular}{lp{0.68\linewidth}}
\toprule
Field & Example \\
\midrule
Instruction & Using the following biomedical data, explain how the biological cascade initiating from Anatomy 'internal carotid artery' reaches Biological Process. \\
\hline
Chosen candidate & This path represents a key hypothesis explaining the biological causality between **'internal carotid artery'** and **'phospholipase C-activating G-protein coupled receptor signaling pathway'**. **[Detailed Mechanism]**: From a clinical perspective, **'internal carotid artery'** forms a biological interaction network with **migraine**, **'migraine'** represents a condition where **Ergotamine** is administered for symptomatic relief or cure, **'Ergotamine'** forms a biological interaction network with **HTR2A**, **'HTR2A'** is involved in **phospholipase C-activating G-protein coupled receptor signaling pathway**, regulating physiological homeostasis. Furthermore ultimately leading to this conclusion. \\
\hline
Rejected candidate 1 & This path represents a key hypothesis explaining the biological causality between **'internal carotid artery'** and **'phospholipase C-activating G-protein coupled receptor signaling pathway'**. **[Detailed Mechanism]**: From a clinical perspective, **'internal carotid artery'** forms a biological interaction network with **migraine**, **'migraine'** represents a condition where **Cyproheptadine** is administered for symptomatic relief or cure, **'Cyproheptadine'** forms a biological interaction network with **HTR2A**, **'HTR2A'** is involved in **phospholipase C-activating G-protein coupled receptor signaling pathway**, regulating physiological homeostasis. Furthermore ultimately leading to this conclusion. \\
\hline
Rejected candidate 2 & This path represents a key hypothesis explaining the biological causality between **'internal carotid artery'** and **'phospholipase C-activating G-protein coupled receptor signaling pathway'**. **[Detailed Mechanism]**: From a clinical perspective, **'internal carotid artery'** forms a biological interaction network with **migraine**, **'migraine'** represents a condition where **Methylergometrine** is administered for symptomatic relief or cure, **'Methylergometrine'** forms a biological interaction network with **HTR2A**, **'HTR2A'** is involved in **phospholipase C-activating G-protein coupled receptor signaling pathway**, regulating physiological homeostasis. Furthermore ultimately leading to this conclusion. \\
\hline
Rejected candidate 3 & This path represents a key hypothesis explaining the biological causality between **'internal carotid artery'** and **'phospholipase C-activating G-protein coupled receptor signaling pathway'**. **[Detailed Mechanism]**: From a clinical perspective, **'internal carotid artery'** forms a biological interaction network with **migraine**, **'migraine'** represents a condition where **Cyclobenzaprine** is administered for symptomatic relief or cure, **'Cyclobenzaprine'** forms a biological interaction network with **HTR2A**, **'HTR2A'** is involved in **phospholipase C-activating G-protein coupled receptor signaling pathway**, regulating physiological homeostasis. Furthermore ultimately leading to this conclusion. \\
\bottomrule
\end{tabular}
\caption{Example of a generated candidate pool.}
\label{tab:generated_candidate_example}
\end{table*}

\begin{table*}[h]
\centering
\small
\begin{tabular}{lp{0.82\linewidth}}
\toprule
Preference label & Candidate path \\
\midrule
Chosen &
internal carotid artery $\rightarrow$
migraine $\rightarrow$
\textbf{Ergotamine} $\rightarrow$
HTR2A $\rightarrow$
phospholipase C-activating GPCR signaling pathway \\
\hline
Rejected 1 &
internal carotid artery $\rightarrow$
migraine $\rightarrow$
\textbf{Cyproheptadine} $\rightarrow$
HTR2A $\rightarrow$
phospholipase C-activating GPCR signaling pathway \\
\hline
Rejected 2 &
internal carotid artery $\rightarrow$
migraine $\rightarrow$
\textbf{Methylergometrine} $\rightarrow$
HTR2A $\rightarrow$
phospholipase C-activating GPCR signaling pathway \\
\hline
Rejected 3 &
internal carotid artery $\rightarrow$
migraine $\rightarrow$
\textbf{Cyclobenzaprine} $\rightarrow$
HTR2A $\rightarrow$
phospholipase C-activating GPCR signaling pathway \\
\bottomrule
\end{tabular}
\caption{Path-only view of the candidate pool in Table~\ref{tab:generated_candidate_example}. All four candidates share the same Anatomy, Disease, Gene, and Biological Process nodes, while differing only in the Compound node (bold), illustrating the entity-slot resolution at which preference learning must operate.}
\label{tab:generated_candidate_path_only}
\end{table*}


\section{Cross-Backbone and Matched-\(K\) Comparison}
\label{app:matched_k}

{We evaluate whether SPA's gains persist under a second model family and whether they can be explained solely by the number of candidates. We repeat the Chemistry-domain comparison using Llama-3.2-3B with the same KARMA-synthesized data and training pipeline. TDPO is evaluated with both \(K=1\) and the matched \(K=4\) setting, while SPA uses \(K=4\).}

\begin{table}[h]
\centering
\small
\resizebox{\columnwidth}{!}{%
\begin{tabular}{lcccc}
\toprule
{Method} &
{MMLU-Chem} &
{MMLU-Pro-Chem} &
{GPQA-Chem} &
{Average} \\
\midrule
{Base} & {0.488} & {0.143} & {0.182} & {0.271} \\
{TDPO (\(K=1\))} & {0.468} & {0.145} & {0.225} & {0.279} \\
{TDPO (\(K=4\))} & {0.495} & {0.153} & {0.236} & {0.295} \\
{SPA (\(K=4\))} & {\textbf{0.553}} & {\textbf{0.179}} & {\textbf{0.284}} & {\textbf{0.339}} \\
\bottomrule
\end{tabular}}
\caption{{Cross-backbone and matched-candidate comparison on the Chemistry domain using Llama-3.2-3B.}}
\label{tab:matched_k}
\end{table}

{Increasing TDPO from \(K=1\) to \(K=4\) improves its average accuracy from 0.279 to 0.295, showing that additional candidates provide useful supervision. Under the same \(K=4\) candidate budget, SPA reaches 0.339 average accuracy and outperforms TDPO on all three benchmarks. The matched comparison indicates that SPA's gain is not explained solely by the number of candidates, while the result on a second model family provides initial evidence of cross-backbone generality.}

\section{{Gradient Analysis}}
\label{app:gradient_analysis}

{We directly examine whether each preference objective concentrates its training signal on the entity-slots where candidates differ. DPO, TDPO, and SPA are trained on the same contrastive candidate pools. DPO and TDPO consume each pool as pairwise comparisons against the same chosen path, while entity annotations are used only post hoc to group tokens for analysis and do not modify either baseline objective.}

{For each method, we measure per-token objective sensitivity separately for chosen-entity tokens, rejected-entity tokens, and shared template tokens. Because absolute gradient magnitudes depend on objective scaling, each group is normalized by the mean sensitivity of template tokens within the same method. The resulting ratios are comparable across token groups within a method, but not as absolute magnitudes across methods. The All Entity / Template ratio is computed as a per-token mean over all entity tokens rather than as the arithmetic average of the chosen and rejected ratios. Results are averaged over nine checkpoints from training steps 100 to 900.}

\begin{table}[h]
\centering
\small
\resizebox{\columnwidth}{!}{%
\begin{tabular}{lccc}
\toprule
{Method} &
{Chosen / Template} &
{Rejected / Template} &
{All Entity / Template} \\
\midrule
{DPO} &
{\(1.00{\pm}0.05\)} &
{\(0.99{\pm}0.06\)} &
{\(0.99{\pm}0.06\)} \\
{TDPO} &
{\(0.80{\pm}0.03\)} &
{\(1.19{\pm}0.04\)} &
{\(1.00{\pm}0.04\)} \\
{SPA} &
{\(\mathbf{2.18{\pm}0.34}\)} &
{\(0.96{\pm}0.12\)} &
{\(\mathbf{1.36{\pm}0.18}\)} \\
\bottomrule
\end{tabular}}
\caption{{Per-token objective sensitivity normalized by the mean template-token sensitivity within each method. Ratios are interpreted only within a method.}}
\label{tab:gradient_analysis}
\end{table}

{DPO assigns nearly uniform sensitivity to chosen entities, rejected entities, and shared template tokens. TDPO places relatively stronger sensitivity on rejected entities but does not concentrate it on the chosen entity-slots. SPA assigns \(2.18\times\) the template-token sensitivity to chosen entity tokens while keeping rejected-entity sensitivity near the template level. This analysis provides direct empirical evidence that SPA routes supervision toward the construction-exposed entity-slots rather than uniformly amplifying all tokens.}

\section{{SPA Loss Component Ablation}}
\label{app:component_ablation}

{We ablate the three components of SPA on the Chemistry domain using Qwen2.5-7B-Instruct. The preference term provides relative supervision between chosen and rejected entity variants, the anchor term increases the absolute likelihood of chosen entities, and the template term preserves the shared language scaffold.}

\begin{table}[h]
\centering
\small
\resizebox{\columnwidth}{!}{%
\begin{tabular}{lcccc}
\toprule
{Configuration} &
{MMLU-Chem} &
{MMLU-Pro-Chem} &
{GPQA-Chem} &
{Average} \\
\midrule
{Base} &
{0.500} &
{0.148} &
{0.161} &
{0.270} \\
{Preference only} &
{0.580} &
{0.182} &
{0.204} &
{0.322} \\
{Preference + Anchor} &
{0.613} &
{0.181} &
{0.161} &
{0.318} \\
{Preference + Template} &
{0.606} &
{\textbf{0.196}} &
{0.161} &
{0.321} \\
{Full SPA} &
{\textbf{0.630}} &
{0.195} &
{\textbf{0.204}} &
{\textbf{0.343}} \\
\bottomrule
\end{tabular}}
\caption{{Ablation of SPA's preference, anchor, and template loss components on the Chemistry domain.}}
\label{tab:component_ablation}
\end{table}

{The preference term alone improves average accuracy from 0.270 to 0.322, showing that slot-level contrastive supervision provides the primary gain. The anchor and template terms improve different evaluation axes when added individually. The full objective achieves the highest average accuracy, the best MMLU-Chem result, a tied-best GPQA-Chem result, and performance within 0.001 of the best MMLU-Pro-Chem result. This pattern indicates that the three components play distinct and complementary roles.}

\end{document}